\documentclass{article}
 
 \pdfoutput=1
    \usepackage[final,nonatbib]{neurips_2020_ml4ad}

\usepackage[table]{xcolor}
\usepackage[utf8]{inputenc} % allow utf-8 input
\usepackage[T1]{fontenc}    % use 8-bit T1 fonts
\usepackage{hyperref}       % hyperlinks
\usepackage{url}            % simple URL typesetting
\usepackage{booktabs}       % professional-quality tables
\usepackage{amsfonts}       % blackboard math symbols
\usepackage{nicefrac}       % compact symbols for 1/2, etc.
\usepackage{microtype}      % microtypography
\usepackage{adjustbox}
\usepackage{multirow}
\usepackage{color,soul}
\usepackage{tabularx}
\definecolor{lightgray}{gray}{0.85}
\usepackage{comment}
\usepackage{epsfig}
\usepackage{epstopdf}
\usepackage{graphics}
\usepackage{caption}
\usepackage{subcaption}
\usepackage{amsmath}
\usepackage{xspace}
\usepackage{float}
\usepackage{graphicx}
\usepackage{anyfontsize}
\usepackage{caption}
\usepackage{enumitem}  % for roman enumerate list
\graphicspath{ {./figs/} }

\title{Investigating the Effect of Sensor Modalities \\in Multi-Sensor Detection-Prediction Models}

\author{%
  Abhishek Mohta, Fang-Chieh Chou, Brian C. Becker,\\ 
  \bf{Carlos Vallespi-Gonzalez, Nemanja Djuric}\\
  Uber Advanced Technologies Group\\
  \texttt{\{amohta, fchou, bbecker, cvallespi, ndjuric\}@uber.com} % 
}

\begin{document}
\maketitle

\begin{abstract}
Detection of surrounding objects and their motion prediction are critical components of a self-driving system. 
Recently proposed models that jointly address these tasks rely on a number of sensors to achieve state-of-the-art performance. 
However, this increases system complexity and may result in a brittle model that overfits to any single sensor modality while ignoring others, leading to reduced generalization. 
We focus on this important problem and analyze the contribution of sensor modalities towards the model performance. 
In addition, we investigate the use of sensor dropout to mitigate the above-mentioned issues, leading to a more robust, better-performing model on real-world driving data.
\end{abstract}

\section{Introduction} 
\label{sect:introduction}

In order to handle traffic situations well and drive safely to their destination, self-driving vehicles (SDVs) are relying on an array of sensors installed on-board \cite{urmson2008self}. 
The set of sensors commonly includes cameras, LiDARs, and radars with a large number of studies analyzing their benefits and disadvantages \cite{fadadu2020multi,shah2020liranet}. 
Multiple sensor modalities are in general leading to improved sensing systems where different sensors complement each other, such as camera helping with detecting objects at longer ranges or radar helping to improve the velocity estimates of vehicles.
However, more sensors being installed also result in a more complex model that may be difficult to manage and maintain, as well as in a brittle system that may exhibit over-reliance on a particular sensor modality (e.g., focusing on a dominant LiDAR sensor during day operations while ignoring other sensors).
In this paper we focus on this critical aspect of the self-driving technology, and analyze the impact of various sensor modalities on the performance of detection and motion prediction approaches. 
For this purpose we consider RV-MultiXNet \cite{fadadu2020multi}, a recently proposed end-to-end system that showed state-of-the-art performance on a number of real-world data sets.

Another important topic of our work is investigation of the use of dropout of different inputs during training.
There are many benefits of such approach, leading to improved generalization performance during online operations. 
In particular, the dropout can help limit the coupling of sensor modalities, for example by preventing the model to mostly rely on radar for vehicle detection and forcing it to also be able to detect vehicles using only LiDAR or only image data.
Moreover, it can also help the autonomous system handle and recover from sensor noise. For instance, when camera is impacted by glare or when sensor input data is dropped due to online latency or hardware issues (such as power interrupts or physical damage to the sensor itself), which can happen during real-world operations. 
Lastly, another important area where sensor dropout can be helpful is in applications that rely on sensor simulation \cite{manivasagam2020lidarsim}. 
The realism gap between simulated and real sensor data is not the same for all sensor modalities. 
As a result, reducing reliance on sensors where this gap is still large can help improve the quality of simulation runs, and lead to a more realistic model performance in simulated environments that is transferable to the real world.

\section{Related Work}
\label{sect:related_work}
Pioneered by the authors of FaF \cite{luo2018fast}, using an end-to-end trained model to jointly perform object detection and motion prediction has become a common technique in autonomous driving.
IntentNet \cite{casas2018intentnet} further incorporates map information and predicts both trajectories and actors' high-level intents. 
LaserFlow \cite{meyer2020laserflow} fuses multi-sweep LiDAR inputs in range-view (RV) representations to perform joint inference.
MultiXNet \cite{djuric2020multixnet} extends IntentNet with second-stage trajectory refinement, joint detection and prediction on multiple classes of actors, as well as multi-modal trajectory prediction. 

Another recent advancement in autonomous driving is learned fusion of multimodal sensor inputs, where the model takes inputs from multiple distinct sensor modalities (e.g., LiDAR, camera, radar), and fuses them to generate the final output.
Continuous fusion \cite{liang2018deep} projects camera features to bird's-eye view (BEV) grid and fuses them with voxelized LiDAR BEV features.
LaserNet++ \cite{meyer2019sensor} performed RV camera-LiDAR fusion for detection. 
\cite{fadadu2020multi} extends MultiXNet with a multi-view BEV+RV architecture and RV camera fusion, while RadarNet \cite{yang2020radarnet} fuses LiDAR and radar inputs.
LiRaNet \cite{shah2020liranet} processes the radar inputs with a spatio-temporal feature extractor and fuses it with LiDAR BEV features to perform joint detection and prediction.
In this work, we use a MultiXNet variant that includes both multi-view camera fusion \cite{fadadu2020multi} and radar fusion \cite{shah2020liranet} as our baseline model.

Dropout \cite{srivastava2014dropout} is a simple yet powerful approach in regularizing deep models, where each output neuron of a layer is dropped with a fixed probability during training. 
Sensor Dropout \cite{liu2017learning} is the most relevant prior work to our method, where the authors randomly dropped a sensor modality completely, instead of performing dropout on individual neurons. 
The authors showed that sensor dropout significantly improves performance of an end-to-end autonomous model when noise is injected into sensor data in simulation. 
In contrast, we applied a similar dropout method to a joint detection-prediction model trained with real-world data instead of using a simulated environment, and performed detailed ablation analysis of the effect of missing sensors and dropout parameters.
\section{Methodology}
\label{sect:methodology}
In this section we present sensor dropout approach to train robust models, capable of jointly detecting and predicting actor motion using LiDAR point clouds, camera RGB images, and radar returns.

\begin{figure}[t]
\includegraphics[width=13cm]{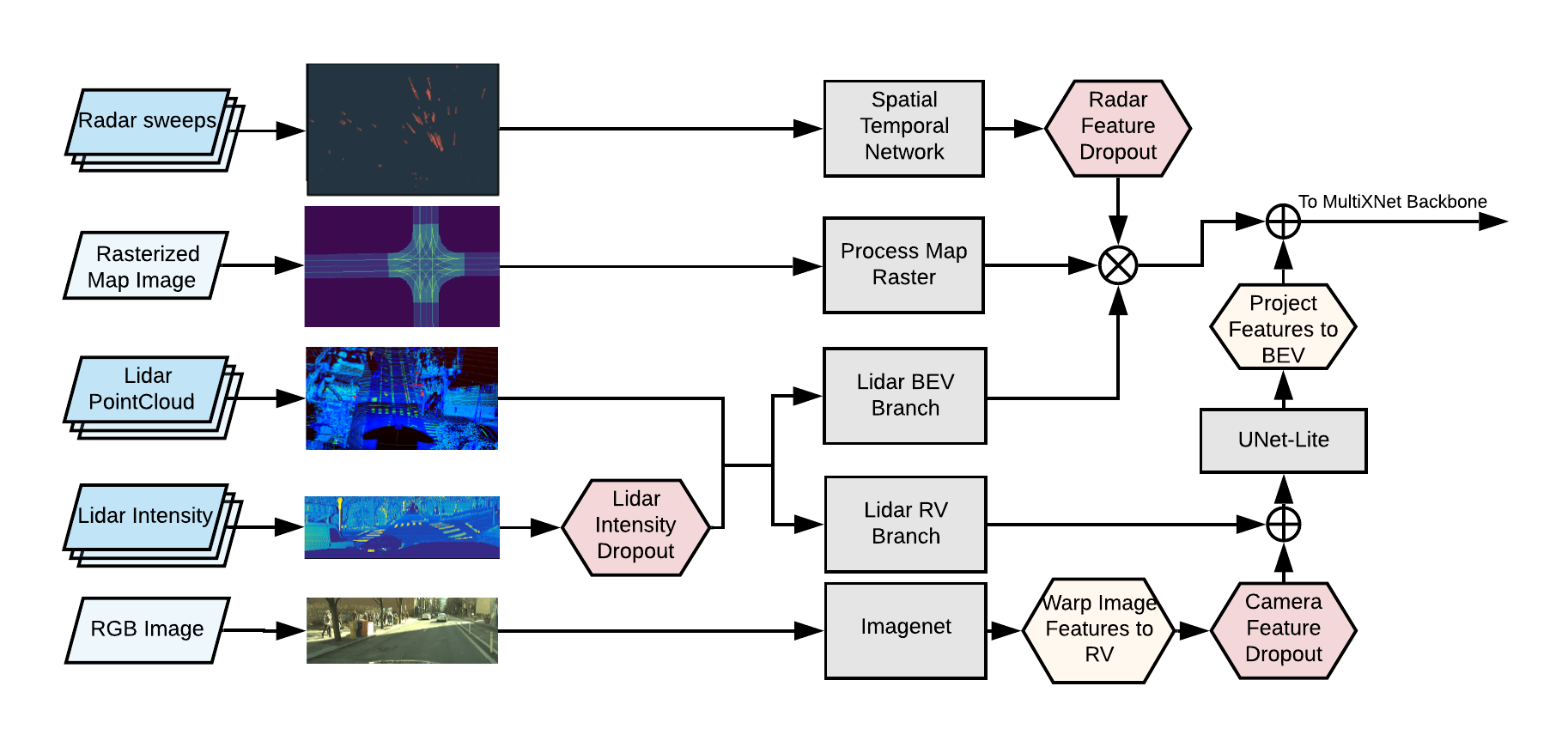}
\centering
\vspace{-0.4cm}
\caption{Network architecture with the proposed sensor dropout}
\vspace{-0.2cm}
\label{figure:arch}
\end{figure}

\textbf{Baseline model:}
The design of our baseline model largely follows the multi-view architecture described in \cite{fadadu2020multi}.
One major modification is that our baseline also includes radar fusion as described in \cite{shah2020liranet}, unlike \cite{fadadu2020multi} that considered only LiDAR and camera inputs.
The model design is illustrated in Figure \ref{figure:arch}, containing an RV branch and a BEV branch. 
In the RV branch the LiDAR points are rasterized into a 2D RV image following feature extraction with a CNN. These features are then fused with the camera image features where U-Net \cite{ronneberger2015u} is used to jointly extract camera and lidar features.
In the BEV branch, $10$ past sweeps of LiDAR points are voxelized onto a BEV grid, then fused with rasterized map channels and radar features extracted by a spatio-temporal network as described in \cite{shah2020liranet}.
Additionally, the RV feature is projected onto BEV and fused with the BEV feature map. 
Finally, the fused feature tensor is processed with additional convolutional layers to output detections and motion predictions for each actor \cite{djuric2020multixnet}.

\textbf{Sensor dropout:}
In this work, we apply sensor dropout to improve the robustness of our model. 
We focus on three sensor modalities used in our baseline model: camera, radar, and LiDAR intensity. 
Note that our model relies on LiDAR point positions to perform multi-view projection and fusion of learned sensor features, therefore it is not viable to drop the LiDAR data completely. 

The sensor dropout involves an approach similar to \cite{srivastava2014dropout}. 
In particular, during training we independently drop each sensor modality with a fixed probability. 
The dropout is performed differently for each sensor modality.
More specifically, for camera and radar we zero out the corresponding final feature vector before sensor fusion is performed.
For LiDAR intensity, we replace the intensity values with a sentinel value, set to the mean LiDAR intensity computed from the training samples.
Figure \ref{figure:arch} illustrates the proposed dropout scheme. 
Note that in the experiments we also explore a variation of sensor dropout for camera, where we zero out input tensors instead of final feature vectors.
\section{Experiments}
\label{sect:experiments}

\textbf{Data and experiment settings:}
We evaluate our method on X17k, a proprietary large-scale autonomous driving data collected in dense urban environment \cite{shah2020liranet}. It contains more than $3$ million frames of samples collected at $10$Hz. We use the same experimental setting as in \cite{fadadu2020multi}, where we use a rectangular BEV input centered at the SDV, with $\textrm{length} = 150$m and $\textrm{width} = 100$m. We use $10$ LiDAR sweeps, $3$ radar sweeps, and the current front camera image to predict $30$ future states.

\setlength\tabcolsep{1.4pt} % default value: 6pt
\begin{table*} 
\normalsize
%\small
\caption{Comparison of AP (\%) and DE ($\text{cm}$) on TCO12 data; improved results shown in bold}
\label{tab:new_eval}
\centering
{
  \fontsize{7.5}{9}\selectfont
  \begin{tabularx}{\textwidth}{c
  >{\centering\arraybackslash}X 
  >{\centering\arraybackslash}X
  >{\centering\arraybackslash}X 
  >{\centering\arraybackslash}X
  >{\centering\arraybackslash}X
  >{\centering\arraybackslash}X 
  >{\centering\arraybackslash}X 
  >{\centering\arraybackslash}X 
  >{\centering\arraybackslash}X
  c} 
       & &\multicolumn{3}{c}{\bf Vehicles} & \multicolumn{3}{c}{\bf Pedestrians} & \multicolumn{3}{c}{\bf Bicyclists}\\
       \cmidrule(lr){3-5} \cmidrule(lr){6-8} \cmidrule(lr){9-11}
       & &\multicolumn{2}{c}{$\textbf{AP}_{0.7}$ $\uparrow$} &  & \multicolumn{2}{c}{$\textbf{AP}_{0.1}$ $\uparrow$} & & \multicolumn{2}{c}{$\textbf{AP}_{0.3}$ $\uparrow$} &  \\
       \cmidrule(lr){3-4} \cmidrule(lr){6-7} \cmidrule(lr){9-10}
       {\bf Method} & {\bf Eval mode} &{\bf Full} & {\bf FOV} & \bf DE$\downarrow$& {\bf Full}&  {\bf FOV } & \bf DE$\downarrow$& {\bf Full}& {\bf FOV}  & \bf DE$\downarrow$\\
    
    \hline
    \rowcolor{lightgray}
    Baseline            &               & 85.8 & 84.7 & 36.0 & 88.1 & 90.3 & 57.5 & 72.9 & 79.1 & 38.0\\
    No camera           &               & 85.9 & 84.6 & 36.2 & 87.7 & 89.0 & 57.5 & 72.4 & 74.5 & 38.0\\
    \rowcolor{lightgray}
    No radar            &               & 85.8 & 84.6 & 37.3 & 87.8 & 90.2 & 57.5 & 73.5 & 77.2 & 36.8\\
    No intensity  &               & 85.8 & 84.8 & 36.0 & 87.3 & 89.9 & 58.3 & 71.6 & 77.4 & 41.2\\
    \rowcolor{lightgray}
    Sensor Dropout       &              & 85.9 & 84.9 & 36.8 & 88.0 & 90.2 & 57.5 & 73.5 & 78.7 & 38.2\\
    \hline
    Baseline            & [-Camera]       & 84.9 & \bf 84.1 & \bf 36.8 & 86.6 & 88.0 & 59.6 & 68.9 & \bf 74.6 & \bf 39.1\\
    \rowcolor{lightgray}
    Sensor Dropout       & [-Camera]       & \bf 85.6 & \bf 84.5 & \bf 37.2 & \bf 87.2 & \bf 88.6 & \bf 58.3 & \bf 71.2 & \bf 74.8 & \bf 38.8\\
    \hline
    Baseline            & [-Radar]       & 81.2 & 83.6 & 41.3 & 86.7 & 89.4 & \bf 57.5 & 70.9 & \bf 77.9 & 44.1\\
    \rowcolor{lightgray}
    Sensor Dropout       & [-Radar]       & \bf 85.3 & \bf 84.7 & \bf 37.7 & \bf 87.8 & \bf 90.1 & \bf 57.3 & \bf 73.3 & \bf 78.4 & \bf 39.6\\
    \hline
    Baseline            & [-Intensity]       & \bf 85.5 & \bf 84.7 & 36.2 & 84.9 & 88.7 & 60.9 & 63.7 & 75.1 & 40.2\\
    \rowcolor{lightgray}
    Sensor Dropout       & [-Intensity]       & \bf 85.8 & \bf 84.7 & 36.8 & \bf 87.0 & \bf 89.6 & \bf 58.6 & \bf 72.2 & \bf 77.9 & \bf 38.4\\
    \hline
\end{tabularx}
}
%\vspace{-0.1cm}
\end{table*}

\textbf{Metrics:}
We follow the same evaluation setting as \cite{fadadu2020multi} for both detection and prediction metrics. 
We report the Average Precision (AP) detection metric, with IoU threshold set to $0.7$, $0.1$, $0.3$ for vehicles, pedestrians, and bicyclists, respectively. 
For prediction metrics we use Displacement Error (DE) at $3$s, with the operating point set at recall of $0.8$. 
As we only use the front camera in our model, we also report detection metrics that only include actors appearing in camera field-of-view (FOV).
To evaluate the model robustness in a degraded setting, we measure the same metrics when a particular sensor modality is missing.
Here we use the same method as for training-time sensor dropout, i.e., zeroing out the final feature of camera/radar and setting LiDAR intensity to a sentinel value in evaluation.
In the following paragraphs we will refer to these evaluation settings as {[-Camera]}, {[-Radar]}, and {[-Intensity]} for metrics with missing camera, radar, and LiDAR intensity inputs, respectively.

\textbf{Results:}
We first performed a sensor-level ablation study, where we trained a baseline model with all sensors, as well as three additional models with each missing one of the three sensors. 
The results are shown in Table \ref{tab:new_eval}. The ablation analysis provides insights into the effect of each sensor input, as well as presenting an upper bound for performance of a robust all-sensor model when a particular sensor modality is missing.
For instance, we can see that without camera the AP metric drops significantly for pedestrians ($1.3\%$) and bicyclists ($4.6\%$) in the camera FOV.
Without radar, the DE metric for vehicles regresses by $1.3$cm, as radar helps provide velocity estimates on the vehicles.
The model with no LiDAR intensity shows $1.5\%$ and $3.2$cm regression in AP and DE for bicyclists, respectively. 

\begin{figure}
\centering
\begin{subfigure}{0.32\textwidth}
  \centering
  \includegraphics[width=\linewidth]{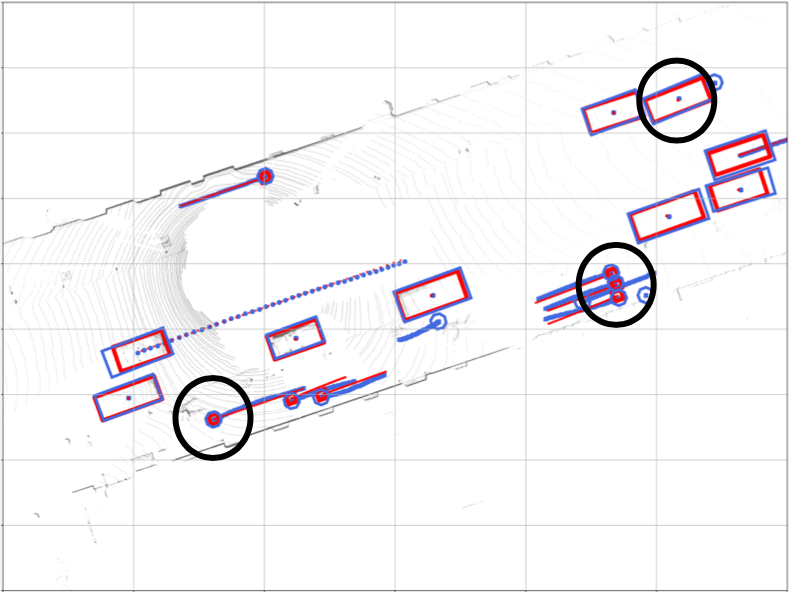}
  \caption{Baseline}
\end{subfigure}%
\begin{subfigure}{0.32\textwidth}
  \centering
  \includegraphics[width=\linewidth]{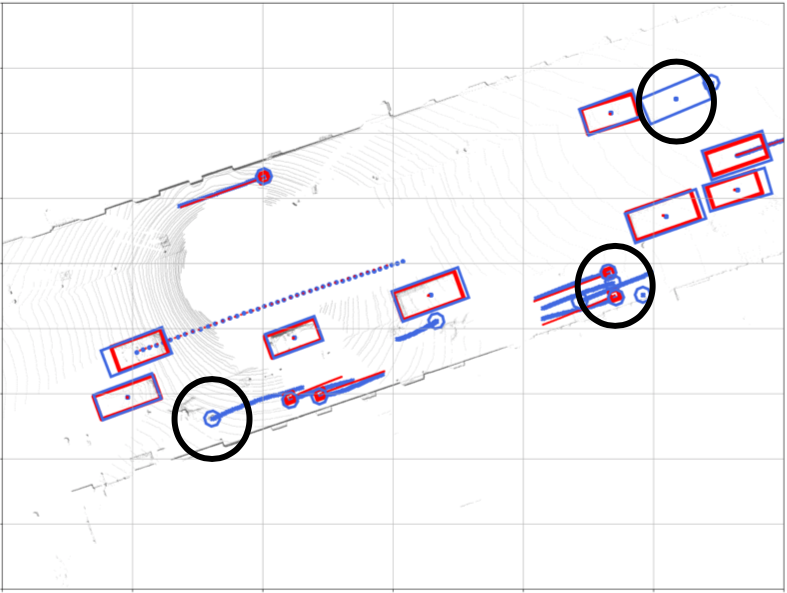}
  \caption{Baseline [-Camera]}
\end{subfigure}%
\begin{subfigure}{0.32\textwidth}
  \centering
  \includegraphics[width=\linewidth]{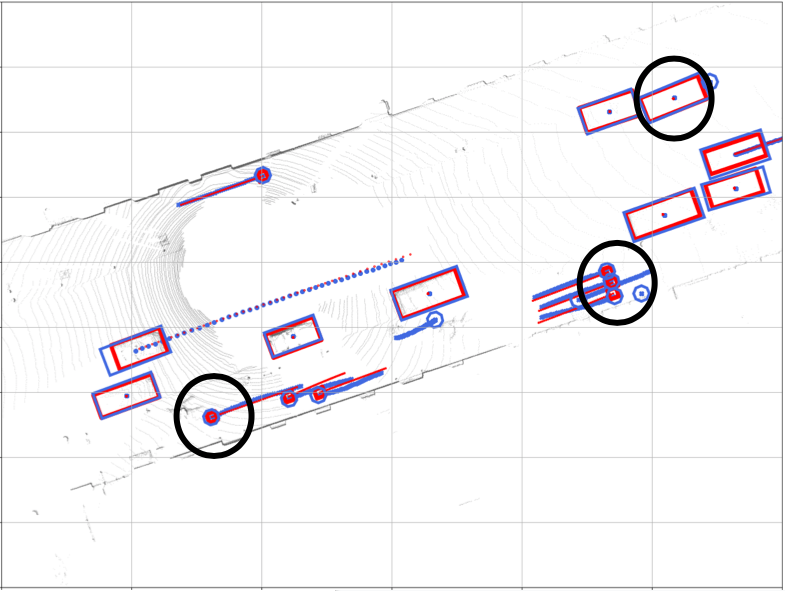}
  \caption{Sensor Dropout [-Camera]}
\end{subfigure}
\caption{Qualitative example showing improved performance with the Sensor Dropout model when the camera input is removed; ground-truth is shown in blue, model detections are shown in red}
\label{fig:qual-2}
\end{figure}

Next, in Table \ref{tab:new_eval} we show how sensor dropout can improve model robustness. 
We first train a sensor dropout model with dropout rate set to $0.2$ for camera and radar and $0.1$ for LiDAR intensity. 
The dropout model shows little change compared to the baseline without dropout, giving comparable metrics. 
However, when evaluated with missing sensor inputs, the dropout model outperforms the baseline for all three missing sensor evaluation settings.
In general, we found that the baseline model metrics regressed significantly when evaluated with missing sensors, while the dropout model gave good results comparable to the performance of sensor ablation models. 
For example, when evaluated without LiDAR intensity (i.e., [-Intensity] rows), the baseline model drops AP for pedestrians by $-2.3\%$ and bicyclists by $-7.9\%$ when compared to \emph{No intensity}, while the sensor dropout model performs comparably ($-0.3\%$ and $+0.6\%$ for pedestrians and bicyclists, respectively).
In Fig. \ref{fig:qual-2} we show a case study of how sensor dropout improves detection in the case of missing sensor.

\begin{figure}
\centering
\begin{subfigure}{0.35\textwidth}
  \centering
  \includegraphics[width=\linewidth]{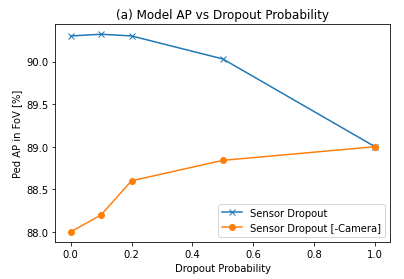}
\end{subfigure}%
\hspace{0.1\textwidth}
\begin{subfigure}{0.35\textwidth}
  \centering
  \includegraphics[width=\linewidth]{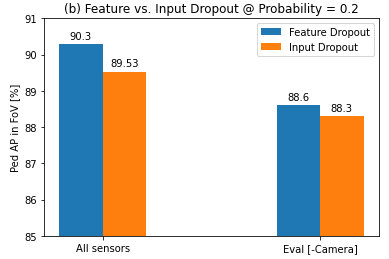}
\end{subfigure}
\vspace{-0.1cm}
\caption{Effect of different settings for camera dropout: (a) sweep over dropout probability values, and (b) feature dropout vs. input dropout}
\label{fig:dropout_params}
\vspace{-0.2cm}
\end{figure}

Lastly, we explore how the dropout probability and different types of dropout algorithm affect the effectiveness of sensor dropout.
In Fig. \ref{fig:dropout_params}a we can see that higher dropout probability leads to lower AP when model is evaluated with all sensors, and improved AP when evaluated with missing sensor, as expected. 
Note that in this work we set the camera dropout probability at $0.2$, which provides a reasonable balance of good performance both with and without missing sensor. 
Fig. \ref{fig:dropout_params}b compares the effectiveness of dropping out the final camera feature (feature dropout) vs. zeroing out the input RGB image (input dropout), where we can see that feature dropout performs better both when evaluated with all sensors and with the missing sensor. 
In input dropout, the model output depends on the learned layers of camera network, while in feature dropout we instead bypass the camera network completely, likely making it more effective by removing the coupling to model's learned weights.

\section{Conclusion}
\label{sect:conclusion}

We performed ablation analyses to understand the effect of each sensor on the performance of a recent multi-sensor detection-prediction model in autonomous driving. 
Furthermore, we applied a sensor dropout scheme to the model, and showed significant improvement of model metrics when evaluated with missing sensor inputs, making the model more robust against sensor failures.
Our proposed dropout method is general and can applied to a variety of autonomous driving models to reduce sensor coupling, to improve robustness against sensor noise, and to make the model metrics more realistic when evaluated on simulated data.

%\pagebreak
\bibliographystyle{abbrv}
{
\small
\bibliography{references}  % .bib
}

\end{document}